\documentclass[pdflatex,sn-mathphys-num]{sn-jnl}

\usepackage{graphicx}%
\usepackage{amsmath,amssymb,amsfonts}%
\usepackage{amsthm}%
\usepackage{mathrsfs}%
\usepackage[title]{appendix}%
\usepackage{xcolor}%
\usepackage{textcomp}%
\usepackage{manyfoot}%
\usepackage{booktabs}%
\usepackage{algorithm}%
\usepackage{algorithmicx}%
\usepackage{algpseudocode}%
\usepackage{listings}%
\usepackage{tabularx} 
\usepackage{booktabs, multirow, rotating, siunitx}
\usepackage[table]{xcolor}
\definecolor{bestrow}{RGB}{230, 242, 255}

\theoremstyle{thmstyleone}%
\theoremstyle{thmstyletwo}%

\theoremstyle{thmstylethree}%

\begin{document}

\title[Article Title]{AnatomicalNets: A Multi-Structure Segmentation and Contour-Based Distance Estimation Pipeline for Clinically Grounded Lung Cancer T-Staging}


\author[1]{\fnm{Saniah} \sur{Kayenat Chowdhury}}\email{kayenat945@gmail.com}

\author[2]{\fnm{Rusab} \sur{Sarmun}}\email{rusabsarmun@gmail.com}
\author*[3]{\fnm{Muhammad} \sur{E. H. Chowdhury}}\email{mchowdhury@qu.edu.qa}
\author[4]{\fnm{Sohaib} \sur{Bassam Zoghoul}}\email{sohaibzoghoul@gmail.com}
\author[5]{\fnm{Israa} \sur{Al-Hashimi}}\email{Ialhashimi@hamad.qa}
\author[6]{\fnm{Adam} \sur{Mushtak}}\email{adamrads94@gmail.com}
\author[7]{\fnm{Amith} \sur{Khandakar}}\email{amitk@qu.edu.qa}


\affil[1]{\orgdiv{Department of Robotics and Mechatronics Engineering}, \orgname{University of Dhaka}, \orgaddress{\city{Dhaka}, \postcode{1000}, \country{Bangladesh}}}

\affil[2]{\orgdiv{Department of Electrical and Electronics Engineering}, \orgname{University of Dhaka}, \orgaddress{\city{Dhaka}, \postcode{1000}, \country{Bangladesh}}}

\affil[3,7]{\orgdiv{Department of Electrical Engineering}, \orgname{Qatar University}, \orgaddress{\city{Doha}, \postcode{2713},\country{Qatar}}}

\affil[4,5]{\orgdiv{Department of Radiology}, \orgname{Hamad Medical Corporation}, \orgaddress{\city{Doha},  \country{Qatar}}}

\affil[6]{\orgdiv{Department of Biomedical Technology}, \orgname{Prince Sattam Bin Abdulaziz University}, \orgaddress{\city{Al-Kharj}, \postcode{11942}  \country{Saudi Arabia}}}


\abstract{Accurate tumor staging in lung cancer is crucial for prognosis and treatment planning and is governed by explicit anatomical criteria under fixed guidelines. However, most existing deep learning approaches treat this spatially structured clinical decision as an uninterpretable image classification problem. Tumor stage depends on predetermined quantitative criteria, including the tumor’s dimensions and its proximity to adjacent anatomical structures, and small variations can alter the staging outcome. To address this gap, we propose AnatomicalNets, a medically grounded multi-stage pipeline that reformulates tumor staging as a measurement and rule-based inference problem rather than a learned mapping. We employ three dedicated encoder–decoder networks to precisely segment the lung parenchyma, tumor, and mediastinum. The diaphragm boundary is estimated via a lung-contour heuristic, while the tumor's largest dimension and its proximity to adjacent structures are computed through a contour-based distance estimation method. These features are passed through a deterministic decision module following the international association for the study of lung cancer guidelines. Evaluated on the Lung-PET-CT-Dx dataset, AnatomicalNets achieves an overall classification accuracy of 91.36\%. We report the per-stage F1-scores of 0.93 (T1), 0.89 (T2), 0.96 (T3), and 0.90 (T4), a critical evaluation aspect often omitted in prior literature. We highlight that the representational bottleneck in prior work lies in feature design rather than classifier capacity. This work establishes a transparent and reliable staging paradigm that bridges the gap between deep learning performance and clinical interpretability.}




\maketitle

\section{Introduction}\label{sec1}
Artificial intelligence has revolutionized medical diagnosis and prognosis, providing AI-driven systems capable of delivering outcomes comparable to expert clinicians ~\cite{leiter2023global, sun2016computer}. In oncology, deep
learning methods have demonstrated remarkable potential for early cancer detection, precise disease classification, and improved treatment strategies~\cite{lakshmanaprabu2019optimal,
wang2022deep, chaunzwa2021deep}. Lung cancer remains a major global health concern and one of the deadliest diseases worldwide, with approximately 2.48 million new cases reported annually~\cite{barta2019global, zhou2024global}. It is the most frequently occurring malignancy in men and the second most commonly diagnosed cancer in women ~\cite{barta2019global}. Therefore, early detection and accurate classification of lung cancer are crucial to improve survival rates for patients~\cite{ambrosini2012pet,
petty2023emerging}. Lung cancer classification follows the Tumor-Node-Metastasis (TNM) staging system, which stratifies disease extent along three independent axes~\cite{osarogiagbon2023international}. The T-stage characterizes the primary tumor's size and extent and is the most spatially complex component of this classification. Under the International Association for the Study of Lung Cancer (IASLC) guidelines~\cite{brierley2025tnm,
goldstraw2016iaslc}, the T-stage is the first component to be assessed, and it categorizes tumors into four main classes: T1 (tumor size $\leq$3\,cm), T2 (tumor $>$ 3cm but $\leq$5\,cm), T3 (tumor $>$5cm but $\leq$7\,cm), and T4 (tumor $>$ 7cm)~\cite{goldstraw2016iaslc, detterbeck2018eighth,
amin2017eighth}. Additionally, T-stage determination is not limited to tumor size alone; it also depends on tumor location and involvement with surrounding anatomical structures such as the mediastinum, thoracic cavity, carina, diaphragm, etc. Moreover, whether the tumor is surrounded only by lung tissue or is nearing the lung walls is a crucial consideration. Thus, the tumor staging becomes an extremely complex clinical task.

\begin{figure}[h]
\centering
\includegraphics[width=0.9\textwidth]{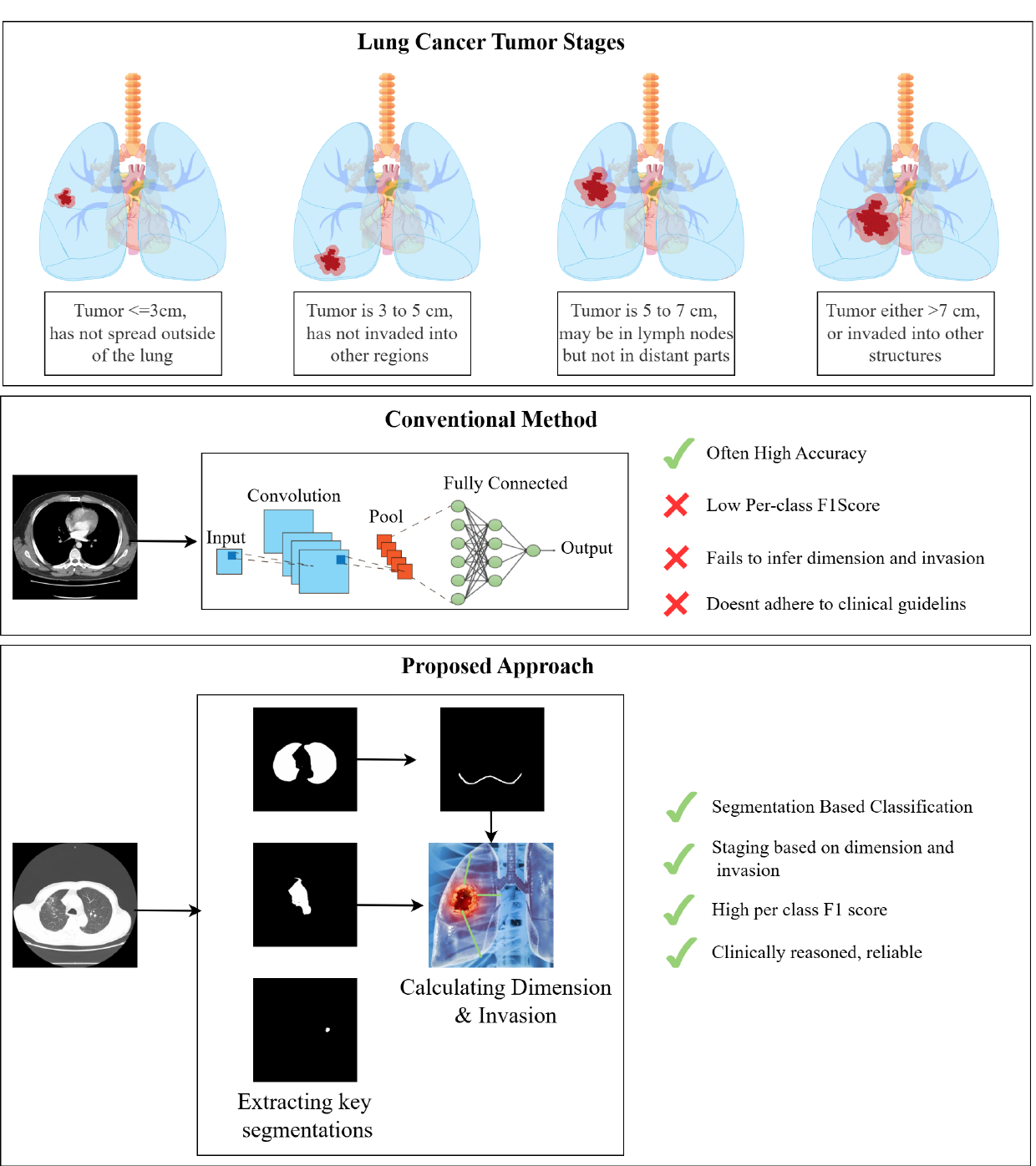}
\caption{Comparing conventional CNN approaches and the proposed segmentation-based pipeline for lung cancer tumor staging, emphasizing the importance of calculating tumor properties for clinical T-staging.}\label{abstract}
\end{figure}

Currently, medical imaging techniques such as Computed Tomography (CT), Positron Emission Tomography (PET), and PET/CT scans are commonly used by experts to diagnose and classify lung cancer, providing detailed anatomical and metabolic information~\cite{ambrosini2012pet, onal2020survival}. However, during staging, clinicians often face limitations in accurately capturing the complex anatomical relationships between the tumor and surrounding structures, leading to potential misclassification. Additionally, the reliance on manual interpretation introduces variability and can be time-consuming. These challenges highlight the need for automation through deep learning methods in order to streamline and enhance the classification process. Existing deep learning approaches for T-stage classification face extreme challenges. Many of these methods treat staging as a pure image classification problem, relying on conventional classifiers to output a stage label. 
This approach has fundamental limitations. A pure Convolutional Neural Network (CNN) classifier does not explicitly explore the anatomical context and quantitative criteria that are essential for T-staging. For example, according to the IASLC guidelines, a tumor with the largest dimension of 6cm is generally labeled as a T3 stage. However, if there is any invasion of mediastinum, diaphragm, esophagus, vertebral body, carina etc., the T-stage is classified as T4 regardless of its size ~\cite{goldstraw2016iaslc}. These conditions are explicitly defined in the IASLC guidelines (described in \ref{sec:staging}), but a generic CNN may struggle with such fine-grained distinctions, failing to recognize multiple organs and the distances between them from a single input~\cite{israel2021factors, chu2024cnn}. Furthermore, CNN-based classification models offer limited interpretability regarding their decision-making criteria, reducing their trustworthiness for borderline cases. Motivated by these limitations, we propose reframing tumor staging as a localization and measurement problem, rather than solely a classification task. This involves distinct segmentation networks tailored for segmenting key regions, including the lungs, mediastinum, diaphragm, and tumor. By explicitly extracting clinically relevant tumor properties (its size, distance to, and invasion of adjacent structures) from these segmentation masks, our method directly adheres to IASLC-defined criteria for T-stage determination. This approach moves beyond end-to-end CNN classifiers and provides an interpretable, reproducible staging pipeline that is clinically relevant. The results achieved via our system highlight the potential of our approach to accurately classify lung cancer T-stage, which could significantly impact clinical decision-making and treatment planning. The comparison in performance between conventional CNN methods and our proposed framework in Table \ref{tab:cnn_comparison} further clarifies the need to incorporate medical context into T-staging.

\section{Literature Review}
\label{sec:lit}

This section provides a comprehensive review of existing literature on the classification of lung cancer subtypes and T-stage classification,
emphasizing the necessity and potential impact of our proposed segmentation-driven
staging framework.

\subsection{Lung Cancer and Nodule Classification}
Multiple studies have shown the potential of CNN-based models to accurately distinguish between
normal and nodular lung tissue on CT scans~\cite{heuvelmans2021lung, hendrix2023deep}.
Moreover, the classification among lung cancer classes has also witnessed massive
success in detecting cancer classes such as ADC, SCLC, SCC, and LCC cancer
types~\cite{adams2018automated, khan2021vgg19, barbouchi2023transformer}. Khan et al.
proposed a VGG-19~\cite{simonyan2014very} based scheme for both segmentation and
classification of lung nodules, achieving a high classification accuracy on two popular
lung tumor datasets~\cite{li2020lungpetctdx} and~\cite{armato2015lidc}. Their use of
combining deep features with hand crafted features benefitted the overall performance
of the network. Wehbe et al.~\cite{wehbe2024enhanced} utilized YOLOv8~\cite{reis2023real}
for lung cancer subtype classification, achieving a mean Average Precision (mAP) score
of 96.8\% at IoU\,=\,0.5. Similarly, Barbouchi et al.~\cite{barbouchi2023transformer}
adopted a Detection Transformer (DeTr)~\cite{dai2021dynamic} model for lung cancer
detection and histologic classification using integrated PET/CT images. It resulted in a
mean IOU of 83\% and an F1-score of 93.66\%. These works are a testament to the success of using deep learning in the classification of lung cancer
classes.

\subsection{T-Stage Classification}

While deep learning has excelled in classification tasks, achieving precise T-staging
based on these spatial criteria remains a significant challenge and an area of research
still relatively unexplored. Barbouchi et al.'s DeTr
transformer~\cite{barbouchi2023transformer} aimed to predict both T-staging and
histologic classification using PET/CT images, classifying T-stages into T1, T2, and
T3/T4. The term `T3/T4' indicates that this work could not differentiate between
T-stage 3 and 4. This work reports a high overall F-1 score (95.63\%) without
highlighting class-specific performances. Moreover, the extent to which the model
explicitly analyses tumor localization and invasion for T-staging is not detailed. Another work by Sathiyamurthy et
al.~\cite{sathiyamurthy2024automated} proposed an automated technique for lung cancer
T-stage detection and classification using an improved U-Net model with an Advanced
Residual Network (ARESNET). Their method involves automated lung nodule mask generation
and utilizes an extended Mobius augmentation technique for data balancing and achieves
an overall accuracy of 94\% across all the classes. This approach incorporates
segmentation, suggesting a reliance on tumor size derived from the segmented mask, but
the explicit analysis of localization and invasion is not thoroughly described. More research to classify T-staging has been done
in~\cite{fan2024tstage, wehbe2024enhanced, zhang2023bayesian}, but these studies all depend solely on traditional CNN models to output a class of T-stage,
without taking the size or location of the tumor into consideration. Therefore, the
question of whether the studies are in accordance with the clinical approach of staging
decisions remains unsolved. The absence of extensive performance breakdown for each
stage class further indicates the necessity for more work. While some studies report
high accuracies in T-stage classification, the methodologies often lack explicit details
on how the models analyze the spatial relationships critical for determining the T
parameter according to the TNM system.

Recognizing the challenges of current deep learning models in T-stage classification,
this research argues for the critical incorporation of the discussed clinically relevant
tumor properties. By developing a model that is sensitive to both the visual patterns
within the tumor and its specific location and size relative to key anatomical
structures, we aim to create a more clinically meaningful and accurate T-stage
classification system. The following are the major contributions presented in our study:

\begin{itemize}
    \item Design and implementation of three dedicated encoder-decoder segmentation
    networks: LungNet, MediNet, and TumorNet for precise segmentation of the lung parenchyma,
    mediastinum, and tumor, respectively, collectively referred to as the AnatomicalNets.

    \item A cross-dataset inference strategy in which the AnatomicalNets are trained
    on multiple publicly available annotated datasets~\cite{simpson2019msd,
    jun2020covid19, medicalsegmentation2020covid, mader2017finding, koitka2024saros}
    and applied to the primary Lung-PET-CT-Dx dataset~\cite{li2020lungpetctdx} as an
    external validation set to generate surrogate segmentation masks for anatomical
    structures that lack manual annotations.

    \item A unique contour-based distance measurement module that extracts the tumor's
    largest dimension and its proximity to the lung walls, mediastinum, and diaphragm
    from the generated segmentation masks.

    \item Implementing an automated pipeline for T-stage classification of lung
    cancer patients based on extracted tumor properties, aligned with IASLC
    guidelines.
    
\end{itemize}

The following sections are structured into four cohesive parts. Section~\ref{sec:method}
offers a deep dive into the models implemented, and the full detail of our methodology. Sections~\ref{sec:experiment} and~\ref{sec:results} demonstrate the
experimental setup and reveal the study's results. The paper culminates in Section~\ref{sec:conclusion}, offering
concluding remarks and insights.


\section{Methodology}
\label{sec:method}

\begin{figure}[h]
\centering
\includegraphics[width=0.9\textwidth]{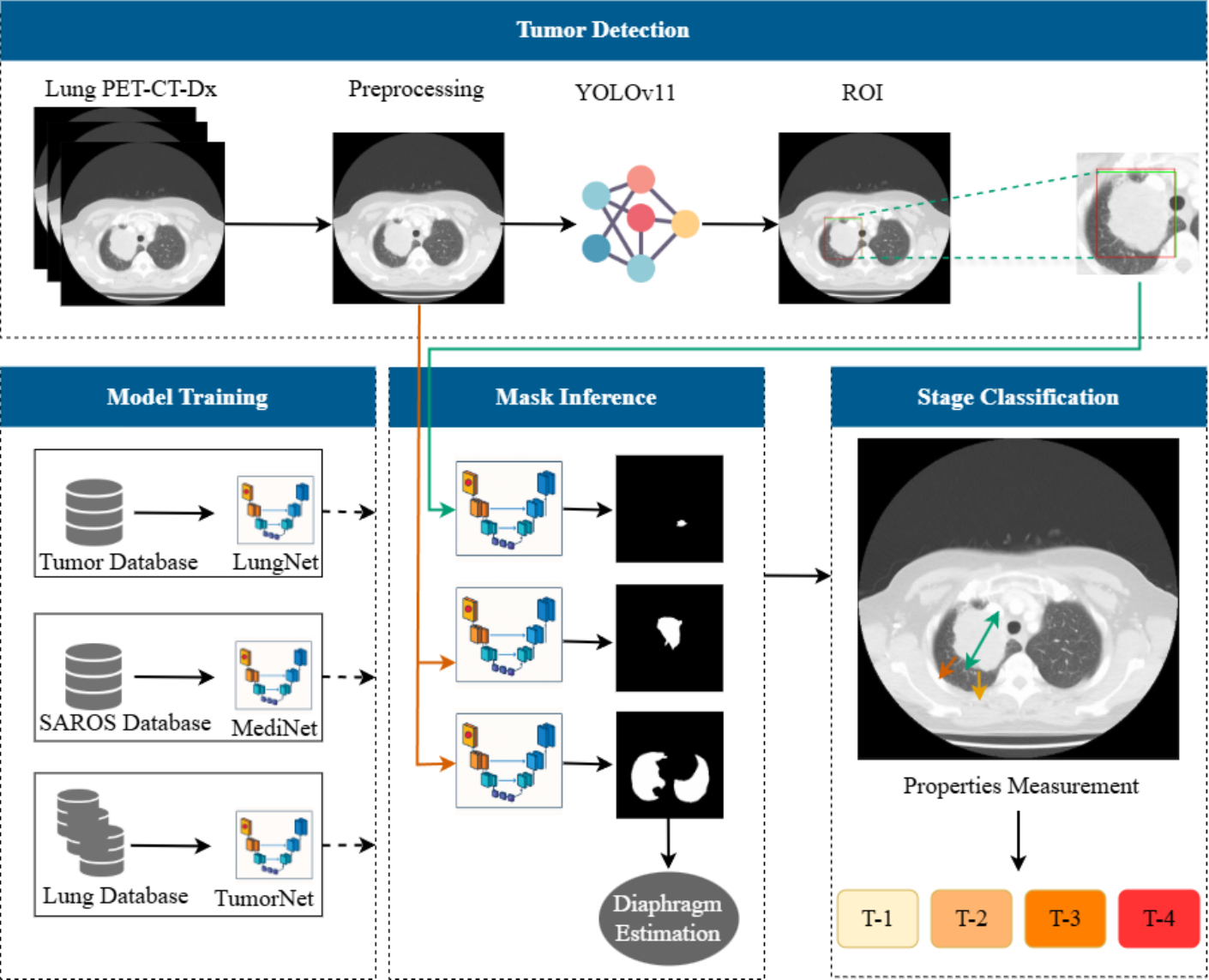}
\caption{Overview of the proposed methodology. The tumor detection module is
    trained on the Lung PET-CT-Dx dataset to localize the tumor region. Corresponding
    databases are used to train the AnatomicalNets. During inference, either the full
    CT image or the detected ROI is used as input to generate segmentation masks. The
    diaphragm position is estimated from lung segmentation masks. Final T-staging is
    performed based on the tumor properties extracted. T-staging: T1 ($\leq$3\,cm and
    surrounded by lung tissue), T2 ($>$3\,cm but $\leq$5\,cm), T3 ($>$5\,cm but
    $\leq$7\,cm), and T4 ($>$7\,cm or invasion into critical structures).}\label{fig:pipeline_overview}
\end{figure}

The proposed methodology for this research involves a multi-stage pipeline designed to
classify the T-stage of lung cancer patients, as illustrated in
Figure~\ref{fig:pipeline_overview}. There are three separate and unique Encoder-Decoder (E-D) CNN
architectures designed and trained in this work, namely:

\begin{itemize}
    \item \textbf{LungNet}: Performs segmentation of the lungs.
    \item \textbf{MediNet}: Performs segmentation of the mediastinum.
    \item \textbf{TumorNet}: Generates precise segmentation masks for the lung tumor.
\end{itemize}

The three models are collectively referred to as AnatomicalNets in this study. The objective of the masks is to extract the necessary tumor size and distance properties for T-stage classification. Due to the lack of a publicly available dataset that simultaneously provides ground truth segmentation masks for all required anatomical structures as well as verified T-stage labels, the three segmentation networks are trained on different datasets~\cite{armato2015lidc, simpson2019msd, jun2020covid19,
medicalsegmentation2020covid, koitka2024saros, mader2017finding} and are applied to the primary dataset Lung PET-CT-Dx~\cite{li2020lungpetctdx} to generate segmentation masks. Additionally, a dedicated detection model is trained on the primary dataset to precisely localize lung tumors within CT slices. Regions of interest (ROI) identified by this detection model are cropped from CT images when they are used as an external
validation set in TumorNet. The last step of the initial stage involves evaluating the tumor's proximity to the
diaphragm using a contour-based distance estimation technique.


In the second stage, the tumor properties required for T-staging are obtained. These properties are calculated using the primary dataset's generated segmentation
masks. The distance is quantified by measuring the maximum difference between the
contours of the respective masks. To obtain the tumor size, the maximum distance
between the contour points of the tumor mask is evaluated. In the final stage, we leverage the obtained tumor properties and perform automated T-stage classification according to the IASLC
conditions. The tumor properties provide information
on the tumor size and whether it is invading other key structures, and T-stage
guidelines are based on such criteria. The following sections of the paper further
detail the data acquisition and preprocessing strategies, model architecture
specifics, and procedures for extracting tumor properties.

\subsection{Datasets Acquisition}
\label{sec:datasets}

This study uses the Lung PET-CT-Dx Dataset~\cite{li2020lungpetctdx} for T-stage
classification as the primary dataset. This dataset contains 220 patients with
individual CT scans, totalling 31,717 CT slices with annotated bounding boxes for
lung tumors. The dataset includes sub-stage labels (e.g., 1b, 2a); since our work
focuses on classifying the major T-stage (T1, T2, T3, T4), three expert radiologists
independently re-evaluated and re-annotated the classifications using the numeric
component only. This resulted in the following stage distribution: T1 (49 patients),
T2 (42 patients), T3 (34 patients), and T4 (95 patients). To train LungNet, we use
three datasets~\cite{jun2020covid19, medicalsegmentation2020covid, mader2017finding},
providing a combined total of 4,616 CT slices. The tumor database consists of one
dataset: the Medical Segmentation Decathlon (MSD)~\cite{simpson2019msd}. Finally,
MediNet is trained on the SAROS database~\cite{koitka2024saros}, a dataset for
whole-body region and organ segmentation in CT imaging, including the mediastinum. These five datasets are referred to as the secondary datasets in this study.
The large amount of collected data across these sources ensures the efficacy and
reliability of the generated masks. A summarisation of all datasets is shown in
Table~\ref{tab:datasets}.

\begin{table}[ht]
\centering
\caption{Summary of the datasets utilized in this work.}\label{tab:datasets}
\begin{tabularx}{\textwidth}{@{}Xcccccc@{}} 
\toprule
\multirow{2}{*}{\textbf{Dataset}}
  & \multirow{2}{*}{\textbf{\shortstack{No. of\\Patients}}}
  & \multirow{2}{*}{\textbf{\shortstack{No. of\\CT Slices}}}
  & \multicolumn{3}{c}{\textbf{Ground Truth Masks}} \\
\cmidrule(lr){4-6}
  & & & \textbf{Lung} & \textbf{Tumor} & \textbf{Mediastinum} \\
\midrule
Lung PET-CT-Dx~\cite{li2020lungpetctdx}
  & 220    & 31,717 & --             & --             & -- \\
COVID-19 CT lung \& infection segmentation~\cite{jun2020covid19}
  & 20     & 3,520  & $\checkmark$   & --             & -- \\
COVID-19 CT segmentation~\cite{medicalsegmentation2020covid}
  & 9      & 829    & $\checkmark$   & --             & -- \\
Finding \& Measuring Lungs in CT Data~\cite{mader2017finding}
  & N/A    & 267    & $\checkmark$   & --             & -- \\
Medical Segmentation Decathlon (MSD)~\cite{simpson2019msd}
  & 64     & 1,224  & --             & $\checkmark$   & -- \\
SAROS~\cite{koitka2024saros}
  & 725    & 5,513  & --             & --             & $\checkmark$ \\
\bottomrule
\end{tabularx}
\end{table}

\subsection{Data Preprocessing, Augmentation, and Split}
\label{sec:preprocessing}

All CT images in the primary and secondary datasets are provided in either DICOM format or NIfTI format. Slices
are set to the lung window (window width: 1400\,HU, window center: $-700$\,HU), and
converted into PNG format images. The image intensities
have been normalized and mapped to pixel values in the range of 0-255.  All PNG images are subsequently resized to $256 \times 256$ pixels for segmentation and detection networks. Contrast-Limited Adaptive Histogram Equalization (CLAHE) is applied to all images to enhance soft-tissue contrast by locally redistributing
pixel intensities without amplifying noise. It is important
to note that the tumor database~\cite{simpson2019msd} includes one additional
preprocessing step. To create highly accurate tumor segmentation masks, the ground
truth masks are superimposed on the CT to extract a rectangular region enclosing the
tumor. Padding is applied for convenience, and the resulting region is cropped from the
CT slice, then resized to $256 \times 256$ pixels before being passed to the tumor
segmentation network. This approach has been beneficial for tumor
segmentation, where it is imperative to capture even the smallest detail.

\begin{figure}[h]
\centering
\includegraphics[width=0.9\textwidth]{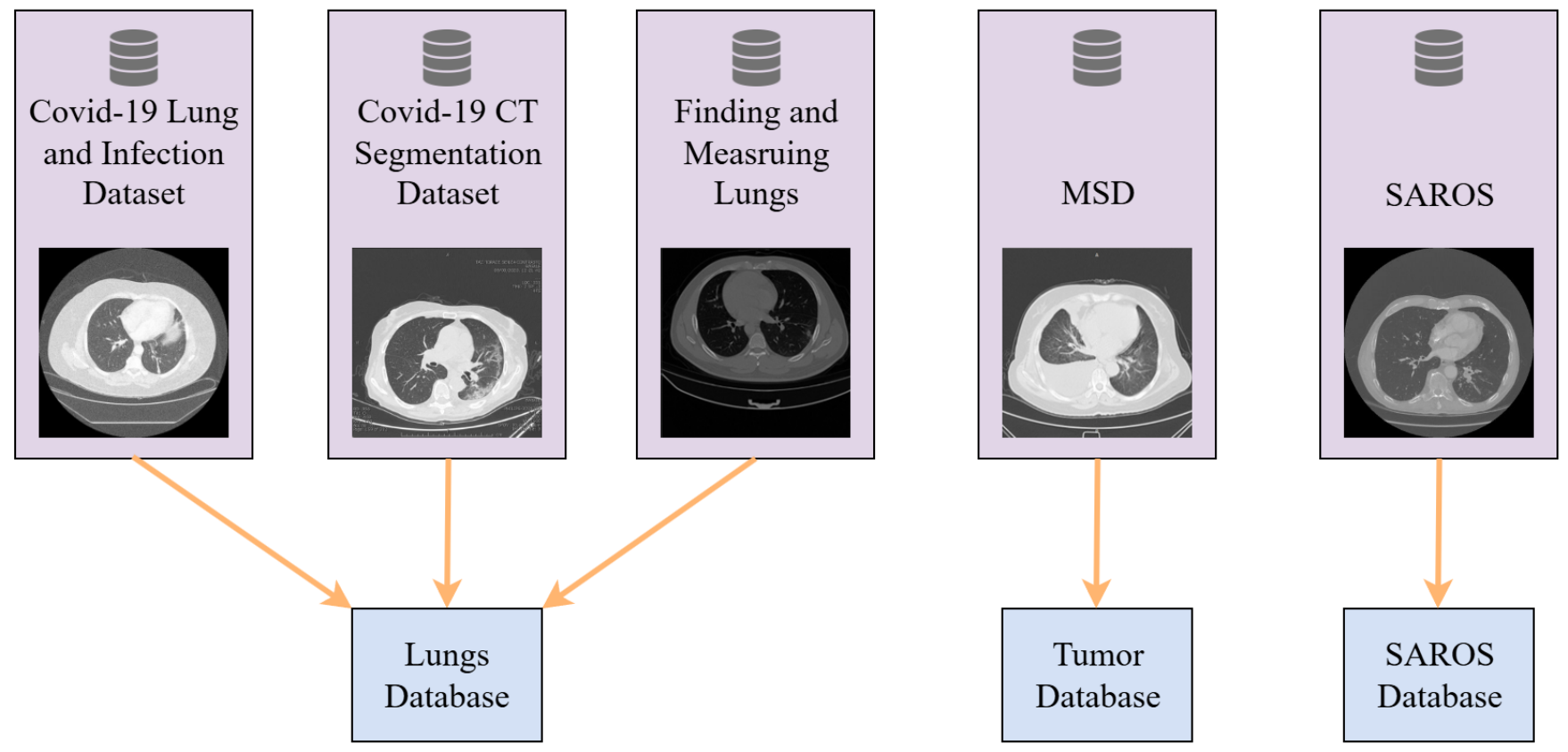}
\caption{Representative CT slices from the datasets incorporated in this study.
    The lung database is compiled from three datasets, while the tumor and SAROS
    databases contain one source each.}\label{fig:ct_samples}
\end{figure}

To ensure robustness, each of the three databases is partitioned into
identical 5-fold cross-validation splits. 80\% of the data is used for training and
20\% for testing. The validation sets contain an additional 20\% drawn from the
training sets. In order to expand the dataset size, we utilize geometric augmentation approaches in the training set. The lung database images are augmented by incorporating horizontal flips and small translation
operations. Due to the relatively smaller number of samples in the tumor database, a
broader range of geometric augmentations is applied to this database, including
horizontal and vertical flips, translation operations, and rotation. Finally, the SAROS
database had geometric augmentations such as horizontal flips and translation operations with a probability of 0.20.

\subsection{Segmentation Models}
\label{sec:segmentation_models}

In this section, we elaborate on the architecture of our three deep learning networks:  LungNet, MediNet, and TumorNet, designed and trained to generate segmentation masks
for the lungs, mediastinum, and tumor respectively. The networks follow a similar base model
architecture, inspired by U-Net~\cite{ronneberger2015unet}. The U-Net
architecture is one of the most impactful architectures
widely adopted for biomedical image segmentation. The U-shaped structure of U-Net
divides the network into two primary paths: the encoder path for contraction and the
decoder path for expansion. As the input progresses through the encoder, the spatial dimensions of the
feature maps decrease, while the number of channels increases, enabling the network to
capture a hierarchical representation of the image. After the encoder, the bottleneck
stage processes the image in a compressed form, allowing the network to interpret the
global context of the image. The decoder path then works to restore the feature map's
original spatial dimensions. U-Net's use of skip connections links feature maps
from each encoder stage to the corresponding decoder stage and enables the integration
of high-level structural information and detailed features, which improves segmentation
accuracy. The U-Net architecture has been further enhanced by researchers using more
complex encoder and decoder
networks~\cite{sarmun2024enhancing, kolhar2021convolutional, lei2021defed}. For
example, ResNet~\cite{he2016deep} and DenseNet~\cite{huang2017densely} are often
chosen as encoders. ResNet is a transformative architecture that introduces residual
connections to train the deeper layers of the network more efficiently and minimize the
problem of vanishing gradients. DenseNet, on the other hand, incorporates dense
connections between layers, allowing each layer to acquire feature maps from all
preceding levels. For the decoder part, the standard U-Net decoder structure is widely
used due to its effectiveness in progressively restoring the resolution of feature
maps. Another popular decoder is the Feature Pyramid Network
(FPN)~\cite{lin2017feature}. FPN utilizes a hierarchical framework comprising encoder
and decoder components arranged in a pyramid-like structure, generating intermediate
segmentation predictions at various spatial resolutions along the decoder pathway.
Ultimately, these intermediate feature maps are resized to match spatial dimensions,
combined, and processed through a convolutional layer with a $3 \times 3$ kernel.
Finally, a Softmax activation function is applied to yield the segmentation mask.
Further advancing the U-Net paradigm, UNet++~\cite{zhou2018unetplusplus} has been
proposed to address some limitations of the original U-Net, particularly in capturing
finer-grained details and improving segmentation accuracy for objects of varying
scales. Unlike the simple skip connections in U-Net, UNet++ redesigns these pathways
to connect the encoder and decoder through a series of nested, dense convolutional
blocks. The dense skip connections enable the aggregation of features from multiple
scales within the decoder, leading to more precise delineation of object boundaries and
better performance on complex segmentation tasks. These approaches have been adopted in this study and the following sections describe each network in detail.

\subsubsection{LungNet}
\label{sec:lungnet}

For accurate T-stage classification, an important criterion is to assess whether the
tumor is completely surrounded by lung tissue, or approaching the lung walls. A tumor
that is fully encased within lung parenchyma and measures less than 3\,cm in its
largest axis is categorised as T1. Hence, LungNet is implemented for producing
high-quality lung segmentation masks, which are to be used to evaluate this condition.
Lung masks are also utilised to approximate the diaphragm position, as diaphragmatic
invasion reclassifies the tumor stage as T4. The E-D network in LungNet uses
DenseNet-121~\cite{huang2017densely} as its encoder. This works efficiently as each
layer in the network is connected to all the other layers during the feed-forward
phase. As a result, each layer accepts all its previous layers as inputs and enhances
feature reusability. For the decoder stage, U-Net's decoder~\cite{ronneberger2015unet}
has shown great performance in the way it increases and restores the resolution of the
encoder output, as well as refines it by using skip connections in our proposed model,
followed by the final $1 \times 1$ convolution. Hence, crucial spatial details and
fine-grained features are maintained during boundary delineation in the lungs.

\subsubsection{MediNet}
\label{sec:medinet}

The second segmentation module in the AnatomicalNets framework is MediNet, developed
and trained for mediastinum segmentation. In T-stage classification, tumor invasion into the mediastinum will directly assign the stage to T4,
irrespective of the tumor size. For this model, we again found
DenseNet-121~\cite{huang2017densely} to perform the best as the encoder. Unlike LungNet, however, the
decoder component is based on the UNet++ architecture~\cite{zhou2018unetplusplus}.
This combination of a DenseNet-121 encoder and a UNet++ decoder allows our model to
leverage the strengths of both architectures: the feature representation power of
DenseNet and the improved feature aggregation of UNet++.

\subsubsection{TumorNet}
\label{sec:tumornet}
The encoder backbone of TumorNet consists of a ResNet-152 model~\cite{he2016deep}, chosen for its
depth and ability to extract detailed hierarchical features. The decoder utilizes a
Feature Pyramid Network (FPN)~\cite{lin2017feature}, designed to effectively
reconstruct segmentation masks by combining multi-scale feature maps generated by the
ResNet-152 encoder. TumorNet operates on cropped tumor ROIs from CT images, enabling
focused segmentation of tumor structures. Deploying a task-specific encoder-decoder
architecture enables us to achieve the highest level of performance, which is evident
in the results section of our experiments.

\subsection{Lung Tumor Detection}
\label{sec:detection}

As illustrated in Figure~\ref{fig:pipeline_overview}, a dedicated lung tumor detection
model based on the YOLOv11 architecture is trained on the primary Lung-PET-CT-Dx
dataset~\cite{li2020lungpetctdx}. This model detects tumors in a CT images using a bounding boxes. These serve as inputs to TumorNet during segmentation inference. The
detection model is an integral part of our pipeline. While training TumorNet, the
tumor region in the annotated tumor database is cropped and padded to generate the
tumor's segmentation mask. However, since there is no ground truth tumor segmentation
mask in the primary dataset, we deploy a detection model to generate tumor ROIs for
each CT slice. The detected bounding boxes are used for mask generation in our
cross-dataset inference strategy. The architecture of YOLOv11 represents a significant enhancement
over previous versions. YOLOv11 incorporates new layers, blocks, and optimizations
that enhance both computational efficiency and detection accuracy. The convolutional
layers assist in gradually decreasing the spatial resolution while increasing the
feature map depth. A special aspect of YOLOv11 is that it uses the C3k2 block
instead of the C2f block. It is a more efficient block based on the
Cross-Stage-Partial (CSP) network. Our lung tumor detection model shows high mean
Average Precision (mAP) scores, providing accurate bounding boxes across CT slices.

\subsection{Mask Inference}
\label{sec:mask_inference}

\begin{figure}[h]
\centering
\includegraphics[width=0.9\textwidth]{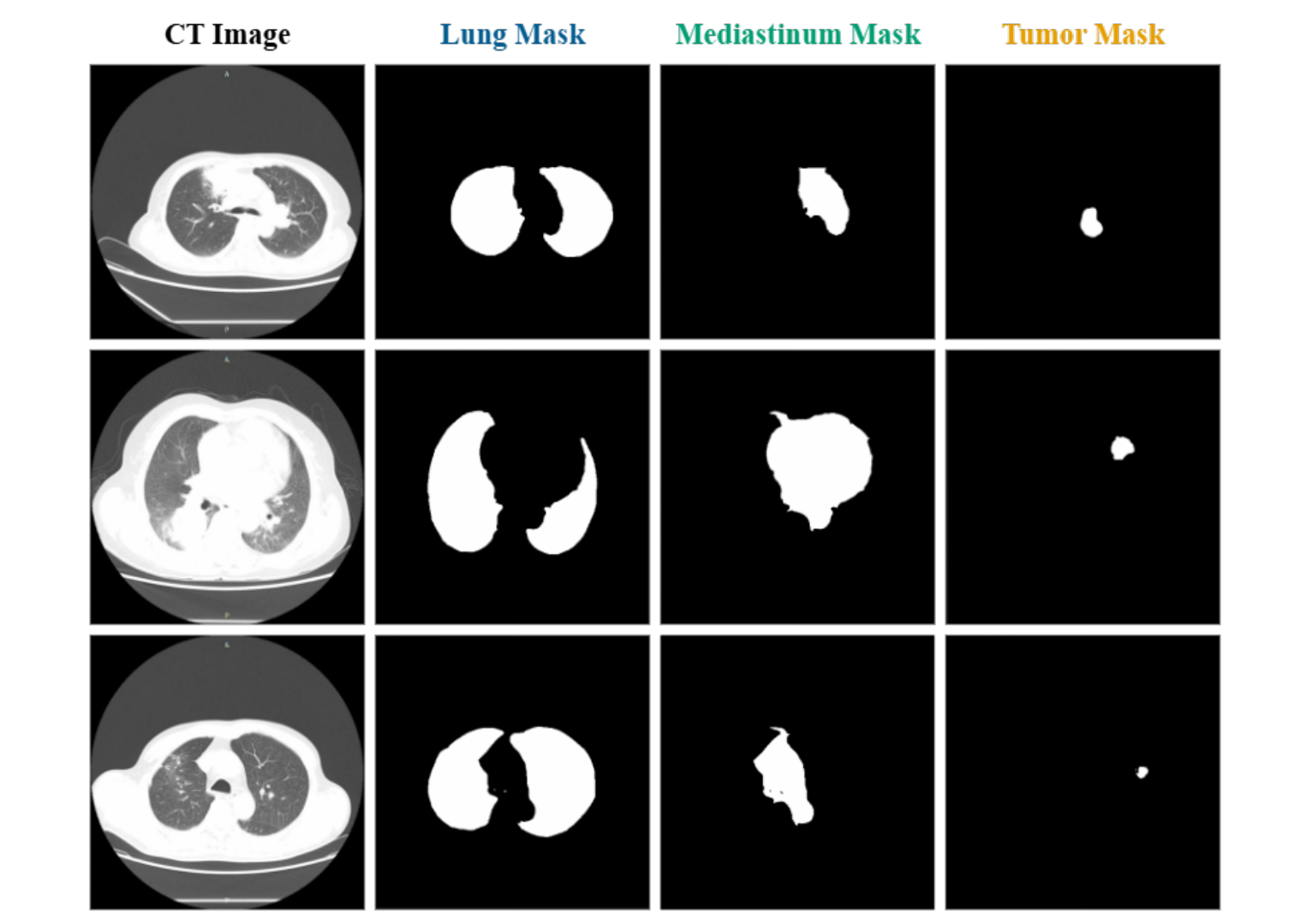}
\caption{CT images (first column) alongside corresponding lung masks (second
    column), mediastinum masks (third column), and tumor masks (fourth column)
    generated through cross-dataset inference.}\label{fig:inferred_masks}
\end{figure}

Since the primary dataset~\cite{li2020lungpetctdx} does not include ground truth masks for
any of the desired anatomical structures, the modules within the AnatomicalNets are trained to be applied to the primary dataset. LungNet, MediNet, and TumorNet
generate ground truth segmentation masks for the lung, mediastinum, and tumor
respectively, by treating Lung-PET-CT-Dx as an external validation set. The inferred
masks serve as surrogate ground truth data, enabling further analysis and research on
the dataset where manual annotations are otherwise unavailable. A randomly selected
subset of the generated masks is reviewed and approved by three expert clinicians for
qualitative assurance. In Figure~\ref{fig:inferred_masks}, sample CT images and
inferred ground truth masks are shown, demonstrating the precision of the generated
lung, mediastinum, and tumor masks.

\subsection{Diaphragm Estimation}
\label{sec:diaphragm}

\begin{figure}[h]
\centering
\includegraphics[width=0.9\textwidth]{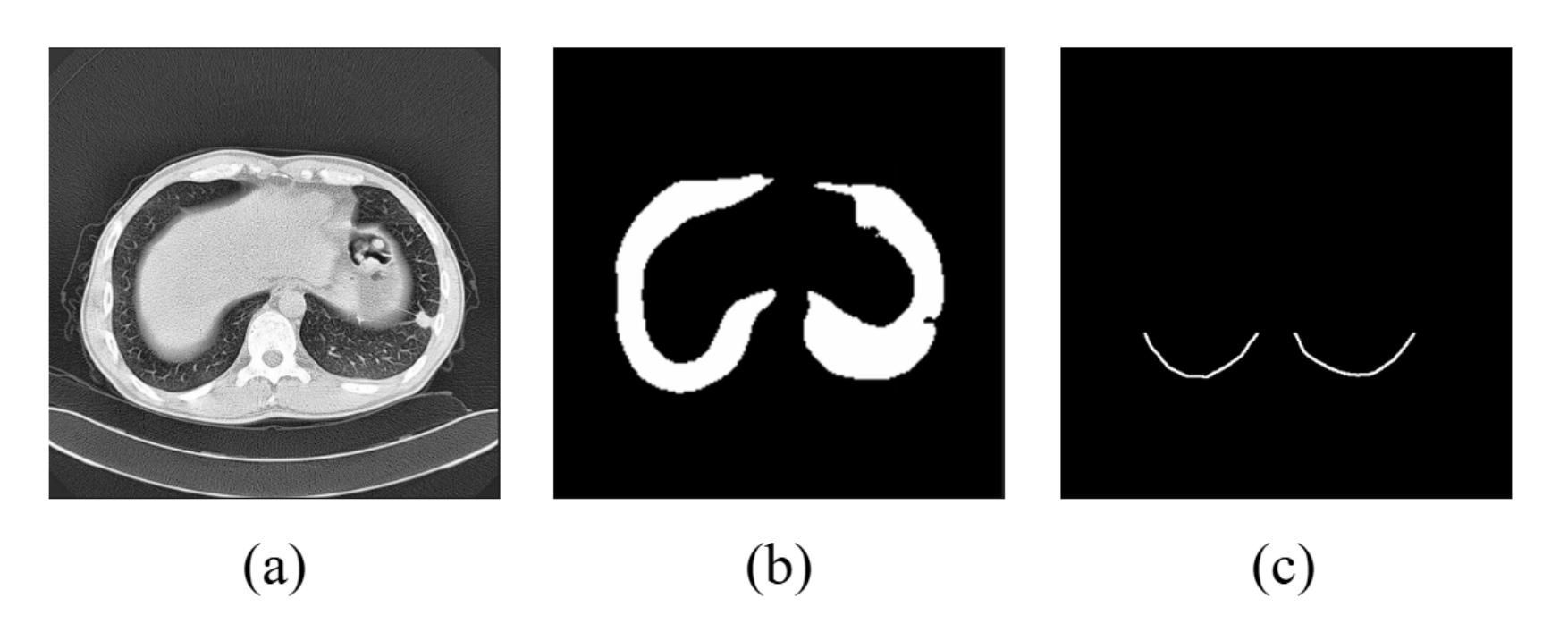}
\caption{(a) A sample CT image from the primary Lung-PET-CT-Dx dataset. (b) The
    corresponding lung segmentation mask generated from the CT image. (c) The
    estimated diaphragm position derived from the lung segmentation mask, used for
    further anatomical analysis.}\label{fig:diaphragm}
\end{figure}

Extracting diaphragmatic invasion is critical for T-stage
classification. However, publicly available datasets specifically annotated for
diaphragm segmentation are limited or unavailable. To overcome this challenge, we
develop a specialized estimation technique leveraging the inferred lung segmentation
masks of the Lung-PET-CT-Dx dataset. Our method involves a pixel-based approach along
the lung mask contours. Specifically, the two lowest pixels at the inferior boundary
of the lung masks are first identified. From these points, an upward region extending
approximately 10\% of the lung mask's height is extracted. The choice of 10\% is not
random but rather obtained from a rigorous trial and error process, indicating that
this method provides the best results. Additionally, insights from studies regarding
the estimation of the diaphragm's position~\cite{suwatanapongched2003variation} aided
in developing this strategy. This strategy enables reliable estimation of the diaphragm's
position without requiring additional segmentation models, offering a practical and
efficient alternative. Figure~\ref{fig:diaphragm} illustrates the step-by-step
procedure for estimating the diaphragm position.

\subsection{Quantitative Measurement of Tumor Properties}
\label{sec:measurement}

The second phase of the proposed pipeline involves computing the tumor properties
essential for classifying T-stage. In clinical practice, to determine the T-stage,
experts evaluate both the size and location of the tumor, particularly in relation
to potential invasion of adjacent structures. To reflect this clinical context, we
calculate the following quantitative metrics from each patient's CT scan:

\begin{itemize}
    \item Maximum tumor dimension.
    \item Distance between the tumor and the lung walls.
    \item Distance between the tumor and the estimated diaphragm.
    \item Distance between the tumor and the mediastinum.
\end{itemize}

\begin{figure}[h]
\centering
\includegraphics[width=0.9\textwidth]{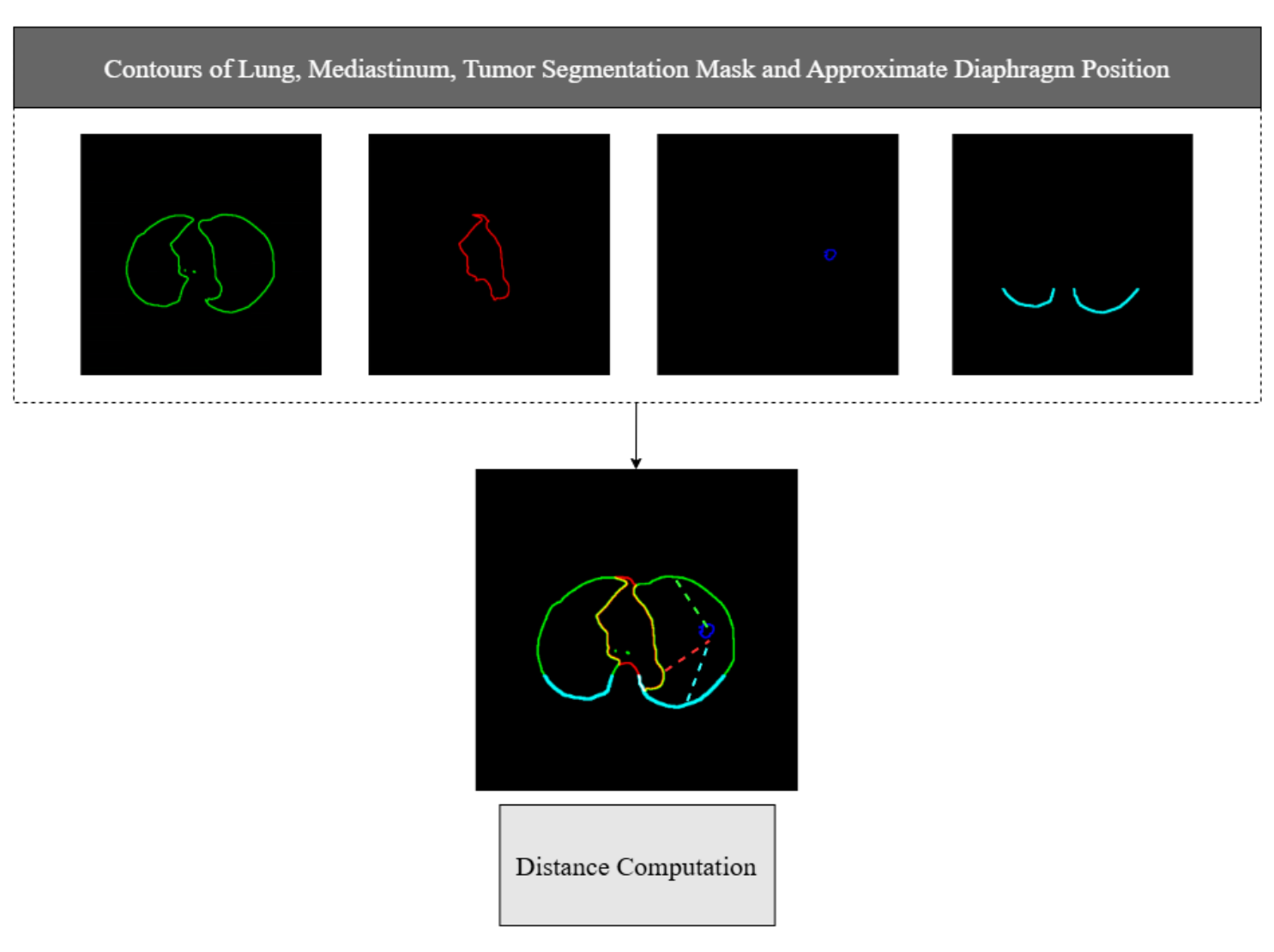}
    \caption{Visualization of the approach used to measure the key distances. The
    first row illustrates the contours obtained from the generated masks. In the
    second row, the contour images are superimposed, and the maximum distances between
    the tumor and key anatomical structures are indicated using a dashed line
    (\textcolor{red}{red} for mediastinum, \textcolor{blue}{blue} for diaphragm,
    \textcolor{green}{green} for lungs).}
    \label{fig:contour_distances}
\end{figure}

These distances are measured by first extracting the contours of the generated ground
truth segmentation masks. Next, we compute the maximum distance between the tumor
contour and the corresponding anatomical region in pixels. Finally, the pixel distance
is multiplied by the pixel spacing of the CT slices to obtain the distance in
practical units. This process ensures that the greatest spatial separation between the
tumor and its neighboring structures is captured, which is crucial for determining
the existence of any invasion. A distance of zero is interpreted as direct invasion of
the corresponding structure. In addition to this, the tumor's largest dimension is
determined. Considering that it can be in the width, height, or depth dimension, two
different approaches have been adopted. First, the contour outline of the tumor
segmentation mask is analyzed to measure the largest Euclidean distance between
contour points, multiplied by the pixel spacing of the CT slice. Additionally, to
estimate the tumor's depth, the number of slices containing visible tumor regions is
multiplied by the slice thickness of the CT scan. Finally, the two dimensions are
compared, and the larger measurement is assigned as the tumor size. These extracted
properties contribute to the overall assessment of the tumor's volume and extent.
Figure~\ref{fig:contour_distances} illustrates the contours from a sample CT image.

\subsection{T-Stage Classification}
\label{sec:staging}

\begin{figure}[h]
\centering
\includegraphics[width=0.9\textwidth]{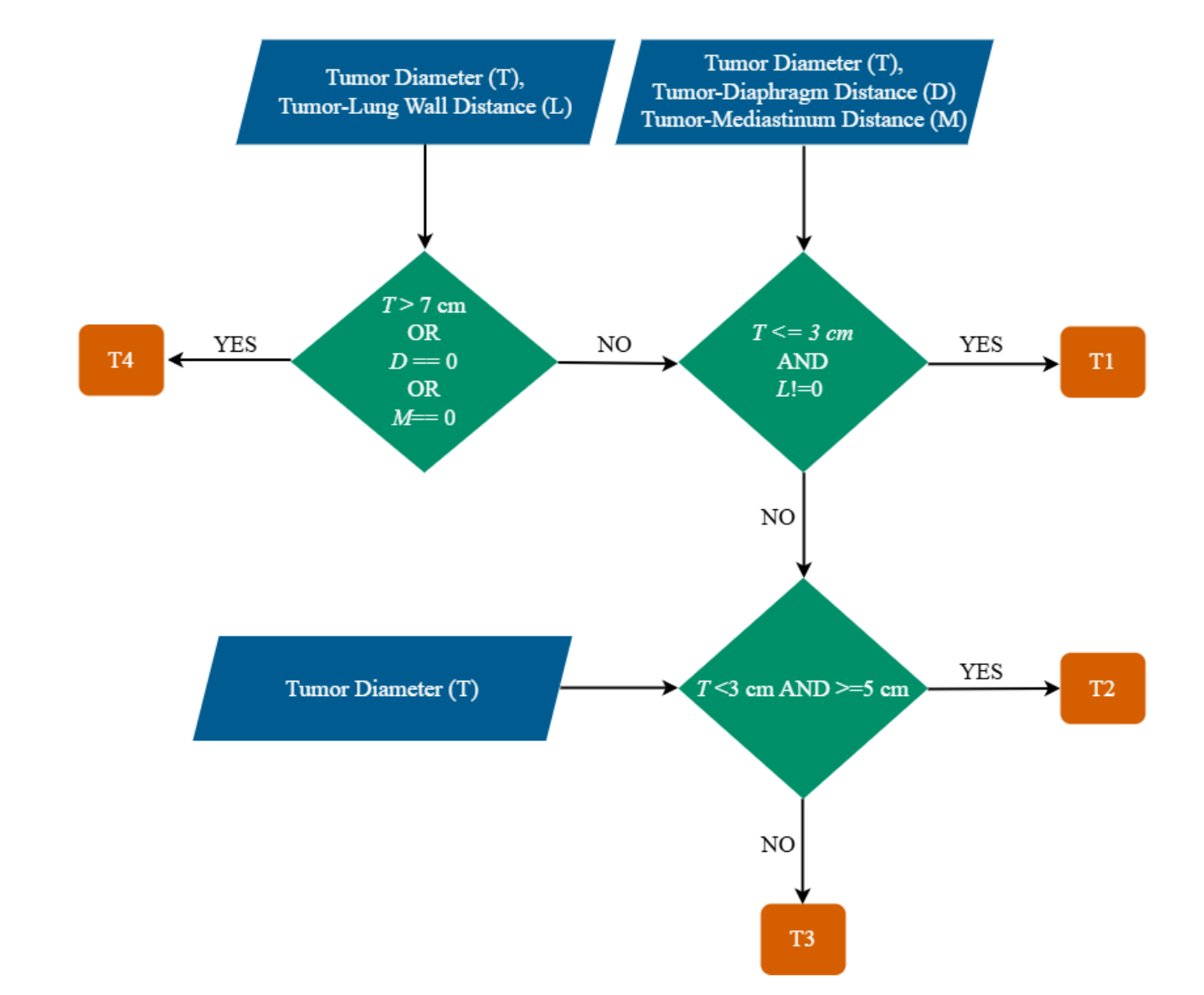}
    \caption{A flowchart describing the step-by-step decision-making process for
    T-stage classification on the basis of tumor size and invasion into other
    anatomical structures.}
    \label{fig:staging_flowchart}
\end{figure}

The final phase of our proposed workflow is the classification of lung cancer T-stage
among patients. The conditions and the decision-making process are detailed in
Figure~\ref{fig:staging_flowchart}. All these conditions are in accordance with the
IASLC guidelines. By adhering to the medical guidelines, our method is not only highly
accurate but also medically reliable and reproducible.

\section{Experiment}
\label{sec:experiment}

This section outlines the experimental framework utilized in our study. We describe
the experimental setup of the architectures in AnatomicalNets, as well as the
detection model using YOLOv11.

\subsection{Experimental Setup}
\label{sec:expsetup}

To achieve the best performance in all of the networks, task-specific training
parameters are used. The training parameters are similar across all architectures
except for the loss function, which has been adapted depending on the network's
purpose. Experiments are conducted on a local workstation equipped with an
Intel\textsuperscript{\textregistered} Core\textsuperscript{\texttrademark}
i9-14900KF CPU at 3.20\,GHz and 32\,GB of RAM, running a x64 architecture.
The software environment comprises Python 3.10.6 and PyTorch 2.6.0+cu118. A number
of state-of-the-art deep learning models are implemented using the Segmentation Models
PyTorch (SMP) package~\cite{iakubovskii2019}. For AnatomicalNets, all advanced
encoders used are initialized with pretrained weights from the ImageNet dataset~\cite{deng2009imagenet}. The optimization of the loss function is performed
using the Adam method, with a constant learning rate ($\alpha = 0.0001$) in the
segmentation networks. In the detection model, Stochastic Gradient Descent (SGD) has
been used as the optimizer. During the training phase, an early stopping criterion is
employed: if no significant improvement in validation loss is observed for 20
consecutive epochs, training is immediately terminated. Hyperparameter and training
parameter details are provided in Tables~\ref{tab:seg_params} and~\ref{tab:det_params}.
\begin{table}[htbp]
\centering
\caption{Segmentation network training parameters.}
\label{tab:seg_params}
\begin{tabularx}{\textwidth}{@{}Xccc@{}}
\toprule
\textbf{Training Parameter} & \textbf{LungNet} & \textbf{MediNet} & \textbf{TumorNet} \\
\midrule
Batch size              & 64        & 64        & 64 \\
Learning Rate           & 1e-4      & 1e-4      & 1e-4 \\
Number of folds         & 5         & 5         & 5 \\
Max epochs              & 100       & 100       & 100 \\
Epoch Patience          & 5         & 5         & 5 \\
Epoch Stopping Criteria & 20        & 20        & 20 \\
Encoder Weight          & ImageNet  & ImageNet  & ImageNet \\
Optimiser               & Adam      & Adam      & Adam \\
Loss Function           & DiceLoss  & DiceLoss  & $0.5 \times \mathcal{L}_{\text{Dice}} + 0.5 \times \mathcal{L}_{\text{Jaccard}}$ \\
Encoder Depth           & 5         & 5         & 5 \\
\bottomrule
\end{tabularx}
\end{table}

\begin{table}[htbp]
\centering
\caption{Tumor detection network (YOLOv11) training parameters.}
\label{tab:det_params}
\begin{tabularx}{\textwidth}{@{}Xl@{}}
\toprule
\textbf{Training Parameter} & \textbf{Value} \\
\midrule
Batch size      & 16 \\
Image Size      & $640 \times 640$ \\
Momentum        & 0.9 \\
Mosaic          & True \\
Label Smoothing & 0 \\
Max epoch       & 100 \\
Optimiser       & SGD \\
\bottomrule
\end{tabularx}
\end{table}


\subsection{Loss Function}
\label{sec:loss}
In LungNet and MediNet, to segment the lung and
mediastinum, DiceLoss has been used. For TumorNet, we optimize our model with an overlap-based loss that combines two complementary measures of mask similarity, Dice and Jaccard, in order to both counteract class imbalance and
sharpen boundary delineation. The Dice component derives from the Dice Similarity
Coefficient (DSC) originally formulated to quantify volumetric overlap in medical
imaging; it naturally ranges between 0 (no overlap) and 1 (perfect agreement), making
it particularly effective when the region of interest occupies only a small fraction of
the scan and models tend to over-predict the background. Since the classical DSC is
non-differentiable, we employ the probabilistic variant, defined for a binary mask as:

\begin{equation}
    \mathcal{L}_{\text{Dice}}
    = 1 - \frac{2\displaystyle\sum_{j=1}^{N} y_j\, p_j + \epsilon}
               {\displaystyle\sum_{j=1}^{N} y_j
              + \displaystyle\sum_{j=1}^{N} p_j + \epsilon}
    \label{eq:diceloss}
\end{equation}

\noindent where $p_j \in [0,1]$ is the model's predicted probability at pixel $j$,
$y_j \in \{0,1\}$ is the corresponding ground-truth label, $N$ is the total number of
pixels, and $\epsilon$ is a small smoothing constant to guard against division by zero.
To further penalize false positives at object boundaries and reinforce overall region
overlap, we augment this with the Jaccard (IoU) loss:

\begin{equation}
    \mathcal{L}_{\text{Jaccard}}
    = 1 - \frac{\displaystyle\sum_{j=1}^{N} y_j\, p_j + \epsilon}
               {\displaystyle\sum_{j=1}^{N} y_j
              + \displaystyle\sum_{j=1}^{N} p_j
              - \displaystyle\sum_{j=1}^{N} y_j\, p_j + \epsilon}
    \label{eq:jaccardloss}
\end{equation}

\noindent By weighting both terms equally, our final overlap loss becomes:

\begin{equation}
    \mathcal{L}_{\text{overlap}}
    = 0.5\,\mathcal{L}_{\text{Dice}} + 0.5\,\mathcal{L}_{\text{Jaccard}}
    \label{eq:overlaploss}
\end{equation}

\noindent This balanced formulation ensures that the network receives strong gradient
signals both where regions should coincide (via the Dice term) and where their
intersection over union must be maximized (via the Jaccard term), leading to robust and
precise tumor segmentation in challenging, low-contrast medical imagery.

\subsection{Evaluation Metrics}
\label{sec:metrics}
In evaluating segmentation performance, this research utilizes a
comprehensive array of measures. In addition to familiar indicators such as precision,
Intersection over Union (IoU), and recall (sensitivity), we also apply the Dice
Similarity Coefficient (DSC), overall accuracy, False Negative Rate (FNR), False
Positive Rate (FPR), and specificity. Among these, IoU and DSC serve as the primary
metrics for this portion of the analysis. The metrics can be defined as follows:

\begin{align}
    \text{IoU}         &= \frac{TP}{TP + FP + FN}                \label{eq:iou} \\[4pt]
    \text{DSC}         &= \frac{2\,TP}{2\,TP + FP + FN}          \label{eq:dsc} \\[4pt]
    \text{Accuracy}    &= \frac{TP + TN}{TP + TN + FP + FN}      \label{eq:acc} \\[4pt]
    \text{Precision}   &= \frac{TP}{TP + FP}                     \label{eq:prec} \\[4pt]
    \text{Sensitivity} &= \frac{TP}{TP + FN}                     \label{eq:sens} \\[4pt]
    \text{Specificity} &= \frac{TN}{TN + FP}                     \label{eq:spec} \\[4pt]
    \text{FNR}         &= \frac{TP \times FN}{TP + FN}           \label{eq:fnr} \\[4pt]
    \text{FPR}         &= \frac{FP}{FP + TN}                     \label{eq:fpr}
\end{align}

\noindent Here, $TP$, $FP$, $FN$, and $TN$ refer to true positive, false positive,
false negative, and true negative, respectively. IoU and DSC are the most impactful
metrics to evaluate the consistency between the ground truth mask and the predicted
mask. There is a slight difference in the way these two metrics are calculated. IoU
evaluates the overlap by determining the ratio of the intersection to the union of the
predicted and ground truth segmentation masks. Conversely, DSC measures overlap by
considering the proportion of the intersection relative to the total aggregate area of
the predicted and ground truth segmentation.

The metrics on which the tumor detection module is evaluated include
precision, recall, F1 score, mean Average Precision (mAP) at an IoU threshold of 0.5
(mAP\textsubscript{50}), and between 0.5 and 0.95 (mAP\textsubscript{50-95}).

The F1 score represents the harmonic mean of precision and recall, providing a
balanced measure between false positives and false negatives:

\begin{equation}
    \text{F1 Score}
    = \frac{2 \times \text{Precision} \times \text{Recall}}
           {\text{Precision} + \text{Recall}}
    \label{eq:f1}
\end{equation}

Additionally, the mAP was computed at two standard thresholds, defined by:

\begin{equation}
    \text{mAP} = \frac{1}{N}\sum_{i=1}^{N} AP_i
    \label{eq:map}
\end{equation}


\section{Results and Discussion}
\label{sec:results}

This section presents an extensive analysis of the performance of the segmentation and
detection experiments conducted in the study, as well as the size and invasion-based
T-stage classification, offering a detailed report of the results. We also compare our
results with results obtained from implementing the most common CNN architectures for
the classification of T-stage.

\subsection{Results of the AnatomicalNets}
\label{sec:results_anatomicalnets}

Table~\ref{tab:segmentation_results} presents an extensive analysis of performance
metrics regarding LungNet, MediNet, and TumorNet. This incorporates a comparison among
five different deep learning networks -- UNet, DenseNet121-UNet, DenseNet121-UNet++,
ResNet50-UNet, and ResNet152-FPN. The assessment criteria comprise Accuracy (Acc),
Dice Similarity Coefficient (DSC), Intersection over Union (IoU), Precision (P),
Sensitivity (SN), Specificity (SP), False Negative Rate (FNR), and False Positive Rate
(FPR). This comparative analysis aims to determine which deep learning network and
preprocessing technique is the most effective in segmenting the target anatomical
regions from CT images. Applying the same five encoder-decoder structures keeps the
consistency among different models and also provides information on which design is
most suitable for the intended task.


\begin{table}[htbp]
\centering
\caption{Model performance comparison for the three anatomical segmentation models
         (mean $\pm$ std across five cross-validation folds).
         Highlighted rows indicate the best-performing configuration selected for
         each network.}
\label{tab:segmentation_results}
\setlength{\tabcolsep}{4pt}
\renewcommand{\arraystretch}{1.2}
\footnotesize

\begin{tabular}{llcccc}
\toprule
\textbf{Model} & \textbf{Network}
  & \textbf{Acc (\%)}
  & \textbf{IoU (\%)}
  & \textbf{DSC (\%)}
  & \textbf{P (\%)} \\
\midrule

\multirow{5}{*}{\textbf{LungNet}}
  & UNet
    & $97.26 \pm 0.31$ & $91.59 \pm 0.52$ & $93.56 \pm 0.44$ & $94.00 \pm 0.61$ \\
  & DenseNet121-UNet++
    & $97.60 \pm 0.28$ & $92.00 \pm 0.48$ & $94.28 \pm 0.39$ & $84.04 \pm 0.87$ \\
  & \cellcolor{bestrow}\textbf{DenseNet121-UNet}
    & \cellcolor{bestrow}$\mathbf{98.96 \pm 0.15}$
    & \cellcolor{bestrow}$\mathbf{96.07 \pm 0.23}$
    & \cellcolor{bestrow}$\mathbf{97.82 \pm 0.18}$
    & \cellcolor{bestrow}$\mathbf{97.27 \pm 0.29}$ \\
  & ResNet50-UNet
    & $97.19 \pm 0.34$ & $92.87 \pm 0.47$ & $95.02 \pm 0.38$ & $94.58 \pm 0.58$ \\
  & ResNet152-FPN
    & $97.88 \pm 0.25$ & $92.79 \pm 0.45$ & $95.98 \pm 0.36$ & $95.64 \pm 0.53$ \\

\midrule

\multirow{5}{*}{\textbf{MediNet}}
  & UNet
    & $97.62 \pm 0.42$ & $88.76 \pm 0.73$ & $91.89 \pm 0.61$ & $90.58 \pm 0.82$ \\
  & \cellcolor{bestrow}\textbf{DenseNet121-UNet++}
    & \cellcolor{bestrow}$\mathbf{98.06 \pm 0.19}$
    & \cellcolor{bestrow}$\mathbf{90.46 \pm 0.38}$
    & \cellcolor{bestrow}$\mathbf{93.39 \pm 0.29}$
    & \cellcolor{bestrow}$\mathbf{91.78 \pm 0.47}$ \\
  & DenseNet121-UNet
    & $97.71 \pm 0.37$ & $89.13 \pm 0.66$ & $92.17 \pm 0.54$ & $90.92 \pm 0.74$ \\
  & ResNet50-UNet
    & $96.78 \pm 0.58$ & $85.92 \pm 0.94$ & $88.47 \pm 0.79$ & $88.21 \pm 1.02$ \\
  & ResNet152-FPN
    & $97.01 \pm 0.51$ & $86.58 \pm 0.87$ & $89.14 \pm 0.72$ & $88.95 \pm 0.94$ \\

\midrule

\multirow{5}{*}{\textbf{TumorNet}}
  & UNet
    & $97.76 \pm 0.38$ & $80.10 \pm 0.84$ & $85.74 \pm 0.71$ & $87.82 \pm 0.93$ \\
  & DenseNet121-UNet++
    & $97.90 \pm 0.33$ & $81.22 \pm 0.77$ & $86.63 \pm 0.64$ & $88.04 \pm 0.85$ \\
  & DenseNet121-UNet
    & $97.88 \pm 0.35$ & $81.76 \pm 0.79$ & $86.45 \pm 0.67$ & $88.39 \pm 0.88$ \\
  & ResNet50-UNet
    & $97.66 \pm 0.41$ & $79.09 \pm 0.91$ & $85.25 \pm 0.76$ & $85.33 \pm 1.01$ \\
  & \cellcolor{bestrow}\textbf{ResNet152-FPN}
    & \cellcolor{bestrow}$\mathbf{97.93 \pm 0.22}$
    & \cellcolor{bestrow}$\mathbf{83.43 \pm 0.58}$
    & \cellcolor{bestrow}$\mathbf{89.68 \pm 0.47}$
    & \cellcolor{bestrow}$\mathbf{89.40 \pm 0.64}$ \\

\bottomrule
\end{tabular}

\vspace{8pt}

\begin{tabular}{llcccc}
\toprule
\textbf{Model} & \textbf{Network}
  & \textbf{SN (\%)}
  & \textbf{SP (\%)}
  & \textbf{FNR (\%)}
  & \textbf{FPR (\%)} \\
\midrule

\multirow{5}{*}{\textbf{LungNet}}
  & UNet
    & $93.76 \pm 0.55$ & $98.30 \pm 0.22$ & $6.24  \pm 0.55$ & $1.70 \pm 0.22$ \\
  & DenseNet121-UNet++
    & $95.64 \pm 0.43$ & $98.59 \pm 0.19$ & $4.36  \pm 0.43$ & $1.41 \pm 0.19$ \\
  & \cellcolor{bestrow}\textbf{DenseNet121-UNet}
    & \cellcolor{bestrow}$\mathbf{96.85 \pm 0.24}$
    & \cellcolor{bestrow}$\mathbf{99.37 \pm 0.09}$
    & \cellcolor{bestrow}$\mathbf{3.15  \pm 0.24}$
    & \cellcolor{bestrow}$\mathbf{0.63 \pm 0.09}$ \\
  & ResNet50-UNet
    & $94.00 \pm 0.51$ & $98.50 \pm 0.21$ & $6.00  \pm 0.51$ & $1.50 \pm 0.21$ \\
  & ResNet152-FPN
    & $94.56 \pm 0.48$ & $98.79 \pm 0.17$ & $5.44  \pm 0.48$ & $1.21 \pm 0.17$ \\

\midrule

\multirow{5}{*}{\textbf{MediNet}}
  & UNet
    & $89.64 \pm 0.77$ & $94.09 \pm 0.55$ & $10.36 \pm 0.77$ & $5.91 \pm 0.55$ \\
  & \cellcolor{bestrow}\textbf{DenseNet121-UNet++}
    & \cellcolor{bestrow}$\mathbf{92.80 \pm 0.41}$
    & \cellcolor{bestrow}$\mathbf{96.63 \pm 0.31}$
    & \cellcolor{bestrow}$\mathbf{7.20  \pm 0.41}$
    & \cellcolor{bestrow}$\mathbf{3.37 \pm 0.31}$ \\
  & DenseNet121-UNet
    & $91.98 \pm 0.68$ & $96.21 \pm 0.44$ & $8.02  \pm 0.68$ & $3.79 \pm 0.44$ \\
  & ResNet50-UNet
    & $88.76 \pm 0.96$ & $95.03 \pm 0.63$ & $11.24 \pm 0.96$ & $4.97 \pm 0.63$ \\
  & ResNet152-FPN
    & $89.39 \pm 0.88$ & $95.41 \pm 0.57$ & $10.61 \pm 0.88$ & $4.59 \pm 0.57$ \\

\midrule

\multirow{5}{*}{\textbf{TumorNet}}
  & UNet
    & $84.56 \pm 0.87$ & $98.55 \pm 0.26$ & $15.44 \pm 0.87$ & $1.45 \pm 0.26$ \\
  & DenseNet121-UNet++
    & $84.26 \pm 0.81$ & $99.51 \pm 0.12$ & $15.74 \pm 0.81$ & $0.49 \pm 0.12$ \\
  & DenseNet121-UNet
    & $84.85 \pm 0.83$ & $99.53 \pm 0.11$ & $15.15 \pm 0.83$ & $0.47 \pm 0.11$ \\
  & ResNet50-UNet
    & $87.33 \pm 0.93$ & $99.52 \pm 0.13$ & $12.67 \pm 0.93$ & $0.48 \pm 0.13$ \\
  & \cellcolor{bestrow}\textbf{ResNet152-FPN}
    & \cellcolor{bestrow}$\mathbf{88.15 \pm 0.61}$
    & \cellcolor{bestrow}$\mathbf{99.55 \pm 0.09}$
    & \cellcolor{bestrow}$\mathbf{11.85 \pm 0.61}$
    & \cellcolor{bestrow}$\mathbf{0.45 \pm 0.09}$ \\

\bottomrule
\end{tabular}

\end{table}

\subsection{Tumor Detection Module}
\label{sec:results_detection}

Table~\ref{tab:detection_results} presents an analysis of performance
metrics regarding the tumor detection model. YOLOv11 has shown the best detection results in terms
of the most important metric mAP\textsubscript{50-95} and other metrics. YOLOv10 and
YOLOv8 have also been utilized to show the difference in performance and to further
emphasise the impact of using YOLOv11. This analysis indicates the efficacy of the YOLOv11
detection model to localize the tumor region from CT images. Similar to the
segmentation models, these are averaged results from the five folds.

\begin{table}[htbp]
\centering
\caption{Model performance comparison for the tumor detection model. Bold row indicates the selected model.}
\label{tab:detection_results}
\begin{tabularx}{\textwidth}{@{}Xcccc@{}} 
\toprule
\textbf{Network} 
  & \textbf{P (\%)} & \textbf{R (\%)} 
  & \textbf{mAP\textsubscript{50} (\%)} 
  & \textbf{mAP\textsubscript{50-95} (\%)} \\
\midrule
YOLOv8    & 85.80          & 83.87          & 86.0          & 45.37 \\
YOLOv10   & 91.00          & 91.00          & 92.3          & 58.6  \\
\textbf{YOLOv11} 
          & \textbf{93.00} & \textbf{92.30} & \textbf{94.3} & \textbf{60.0} \\
\bottomrule
\end{tabularx}
\end{table}

\subsection{T-Stage Classification}
\label{sec:results_classification}

\begin{figure}[h]
\centering
\includegraphics[width=0.9\textwidth]{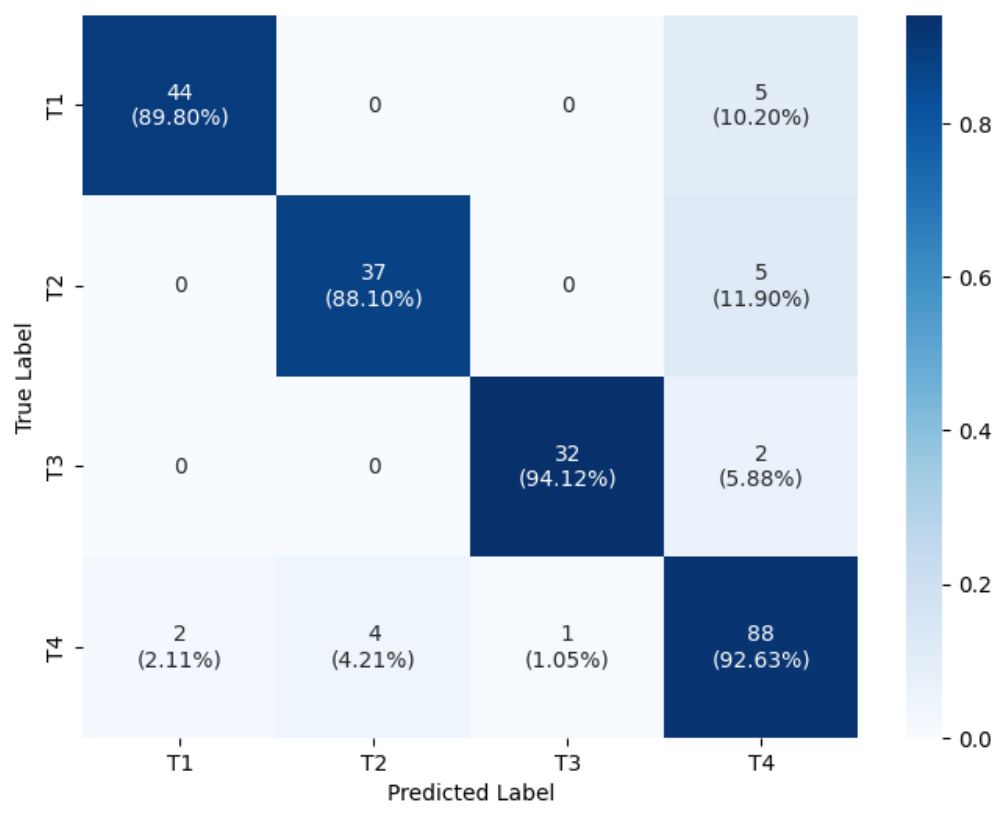}
    \caption{Confusion matrix illustrating the performance of the proposed novel
    T-stage classification approach.}
    \label{fig:confusion_matrix}
\end{figure}

The core objective of this study is accurate classification of the T-stage of lung
cancer patients. To keep this classification process aligned with clinically
established guidelines, we opt for a method that performs classification based on
tumor size and its invasion into other significant structures. This novel approach
achieves high metrics across all significant evaluation criteria. The classification is done according to the procedure described in Section~\ref{sec:staging}. The classification report is summarized in
Table~\ref{tab:classification_results} and a confusion matrix is provided in
Figure~\ref{fig:confusion_matrix}. The proposed methodology achieves an overall
accuracy of 91.36\%. The detailed breakdown of precision, recall, and F1 score for
each T-stage (T1, T2, T3, T4) further highlights the efficacy of the methodology. All
classes achieve high F1 scores, indicating the robustness of the model in classifying
the stages. The confusion matrix illustrates a pattern in the cases the approach
failed to make the correct predictions. To further validate the proposed methodology
over traditional image classification networks, a comparison in performance is
demonstrated in Table~\ref{tab:cnn_comparison} between our approach and the most
widely adopted CNN architectures. These differences further indicate the credibility
and necessity of the adopted approach.

\begin{table}[htbp]
\centering
\caption{ Classification results of lung cancer T-stage.}
\label{tab:classification_results}
\begin{tabularx}{\textwidth}{@{}Xcccc@{}} 
\toprule
\textbf{T-Stage}
  & \textbf{Acc (\%)} & \textbf{P (\%)} & \textbf{R (\%)} & \textbf{F1 Score} \\
\midrule
T1 & \multirow{4}{*}{91.36} & 96.0 & 90.0 & 0.93 \\
T2 &                         & 90.0 & 88.3 & 0.89 \\
T3 &                         & 97.0 & 96.0 & 0.96 \\
T4 &                         & 88.0 & 90.0 & 0.90 \\
\bottomrule
\end{tabularx}
\end{table}

\begin{table}[htbp]
\centering
\caption{Comparison of performance with traditional CNN approaches.}
\label{tab:cnn_comparison}
\begin{tabularx}{\textwidth}{@{}Xcccc@{}} 
\toprule
\textbf{Network}
  & \textbf{Acc (\%)} & \textbf{P (\%)}
  & \textbf{R (\%)}   & \textbf{F1 Score (\%)} \\
\midrule
ResNet152~\cite{he2016deep}             & 37.29 & 39.04 & 37.29 & 36.71 \\
DenseNet121~\cite{huang2017densely}     & 40.35 & 39.80 & 39.80 & 39.81 \\
Swin Transformer                        & 39.28 & 40.08 & 39.28 & 39.04 \\
\midrule
\textbf{Proposed Model}
  & \textbf{91.36} & \textbf{96.0}
  & \textbf{90.0}  & \textbf{92.00} \\
\bottomrule
\end{tabularx}
\end{table}

Table~\ref{tab:classification_results} portrays the reliability of the
proposed framework, highlighting consistently high F1 scores ranging from 0.89 to 0.96
across different tumor stage categories. This indicates the model's strong capability
in accurately distinguishing each class. Table~\ref{tab:cnn_comparison} provides a
comprehensive comparative analysis with existing CNN-based approaches. These methods
exhibit a maximum average F1 score of 39.81\% and an accuracy of 40.35\%,
significantly inferior to our results. These findings support the approach of adopting
a clinically informed framework to tumor staging.

\subsection{A Deeper Look into the Wrong Predictions}
\label{sec:wrong_predictions}

It is observable from Table~\ref{tab:classification_results} and
Figure~\ref{fig:confusion_matrix} that stage T4 has the lowest precision and the
second lowest F1 score among all the stages. This can be explained using clinical
rationale. Key points to consider include:

\begin{itemize}

    \item \textbf{Multifocal Disease as a T4 Criterion.} A key criterion for T4
    classification in lung cancer is the presence of separate tumor nodules in a
    different ipsilateral lobe than the primary tumor~\cite{goldstraw2016iaslc,
    detterbeck2018eighth}. A patient with such tumor nodules is classified as T4,
    regardless of the individual nodule sizes or the extent of invasion of contiguous
    structures by any single nodule. In our proposed pipeline, which primarily assesses
    individual nodule characteristics (size) and their distances to anatomical
    landmarks, a patient with, for instance, a 2\,cm nodule in the upper lobe and a
    1.5\,cm nodule in the lower lobe could be erroneously classified as T1 or T2 based
    solely on these individual measurements. This limitation significantly contributes
    to the observed misclassification of true T4 cases as earlier stages.

    \item \textbf{Limited Scope of Invaded Structures in the T4 Definition.} The
    misclassification of T4 is also significantly influenced by the limited range of
    invaded structures considered by our model. According to the AJCC 8th Edition TNM
    classification, a primary lung tumor is classified as T4 if it directly invades
    any of the following structures: diaphragm, mediastinum, heart, great vessels
    (e.g., aorta, superior vena cava, pulmonary artery/veins), trachea, recurrent
    laryngeal nerve, esophagus, vertebral body, or carina~\cite{goldstraw2016iaslc,
    detterbeck2018eighth, amin2017eighth}. Our model's features are restricted to
    distance measurements from only the diaphragm, mediastinum, and the lung walls.
    While our method presents a novel approach and is among the first works to address
    staging in a clinically relevant manner using such features, this narrow focus
    represents a significant limitation causing these misclassifications.

\end{itemize}

These points collectively highlight the inherent complexity of diagnosing the T4 stage
and directly indicate the limitations of this study's current feature set.
Table ~\ref{tab:misclassified} illustrates a few examples of misclassified T4
cases.

\newcolumntype{L}{>{\raggedright\arraybackslash}p{3.5cm}} 

\begin{table}[htbp]
\centering
\caption{Examples of misclassified CT images in the T4 stage.}
\label{tab:misclassified}
\begin{tabularx}{\textwidth}{@{}X L c c @{}} 
\toprule
\textbf{Patient ID} & \textbf{Tumor Properties (cm)} & \textbf{Predicted Stage} & \textbf{Ground Truth} \\
\midrule
51 & T=4.2, L=0.93, M=0.71, D=1.84 & T2 & T4  \\
\cmidrule(r){1-4}
93 & T=6.3, L=0.58, M=0.86, D=2.2 & T3 & T4  \\
\cmidrule(r){1-4}
134 & T=2.1, L=1.5, M=2.4, D=3.2 & T1 & T4  \\
\bottomrule
\end{tabularx}
\end{table}


\subsection{Limitations \& Future Work}
\label{sec:limitations}

In this study, we have demonstrated a novel hybrid framework that integrates
deep-learning-based segmentation of key thoracic structures with intuitive, rule-based
measurements, yielding interpretable T-stage classifications at high accuracy. However,
there are some limitations of this approach. Addressing the limitations could serve as
motivation for future work in this domain of research. Our diaphragm boundary relies
on a simple heuristic using the lowest 10\% of lung mask points, due to the lack of
datasets related to diaphragm segmentation. It may misalign in atypical anatomies.
Also, our evaluation remains confined to a single 2D CT dataset. By focusing solely on
two-dimensional slices, we may overlook volumetric tumor characteristics that could
refine staging. Furthermore, according to the IASLC
guidelines~\cite{goldstraw2016iaslc, detterbeck2018eighth}, the staging of lung tumors
also depends on additional structures visible in the CT scan, such as the trachea,
esophagus, and carina. In future work, we plan to curate or annotate a true diaphragm
segmentation set to replace our heuristic, extend the pipeline to full 3D segmentation
for more accurate tumor depth assessment, and perform segmentation of additional
regions to broaden validation across multi-center cohorts. Finally, we aim to
expand our approach to complete TNM staging by integrating lymph node and metastasis
detection models to further enhance clinical applicability.


\section{Conclusion}
\label{sec:conclusion}

This study proposed a novel, transparent, clinically informed framework for lung cancer
T-stage classification that synergizes deep-learning-based segmentation with rule-based
measurement protocols. By explicitly modeling anatomical context and adhering to IASLC
criteria, our method attains 91.36\% accuracy and robust per-stage performance across four tumor stages. The
approach bridges the gap between black-box classifiers and clinical practice, offering
interpretable outputs that can readily integrate into radiological workflows. Future
enhancements, such as refined diaphragm segmentation, 3D volumetric modeling, and
expansion to full TNM staging, will further elevate its utility and generalizability
in real-world settings.



\backmatter








\section*{Declarations}
\begin{itemize}
\item \textbf{Funding} This study did not receive any funding
\item \textbf{Conflict of interest} No conflict of interest to declare
\item \textbf{Data availability } The datasets used in this study are publicly available
\end{itemize}











\bibliography{sn-bibliography}

\end{document}